%% file: paper.tex
\documentclass[runningheads]{llncs}
\usepackage{graphicx}

\newif\ifanonymous

\usepackage[utf8]{inputenc}  
\usepackage[T1]{fontenc}     
\usepackage{graphicx}        
\usepackage{orcidlink}
\usepackage{adjustbox}
\usepackage{hyperref}

\begin{document}

\title{Enabling Low-Resource Language Retrieval: Establishing Baselines for Urdu MS MARCO}

\titlerunning{Enabling Low-Resource Language Retrieval} 

\ifanonymous
    \author{Anonymous Author(s)}
    \authorrunning{Anonymous}
    \institute{Anonymous Institution(s)}
\else
    \author{Umer Butt\inst{1,2} \orcidlink{0009-0004-2547-9334} \and
    Stalin Varanasi\inst{1,2} \orcidlink{0009-0006-1051-1447} \and
    Günter Neumann\inst{1,2} \orcidlink{0000-0002-8929-2729}}

    \authorrunning{U. Butt et al.}
    
    \institute{University of Saarland, Germany \\
    German Research Center for Artificial Intelligence (DFKI), Germany \\
    \email{\{mubu01, stalin.varanasi, guenter.neumann\}@dfki.de}
    }
\fi

\maketitle

\begin{abstract}
\input{abstract}
\end{abstract}

\section{Introduction}
\input{introduction}

\section{Related Work}
\input{related_work}

\section{Dataset Creation}
\input{dataset}

\section{Experimental Setup}
\input{setup}

\section{Results and Discussion}
\input{results}

\section{Future Work and Conclusion}
\input{conclusion}

\section{Acknowledgement}
\input{acknowledgement}

\newpage

\bibliographystyle{splncs04}  
\bibliography{references}     

\end{document}

%% file: abstract.tex
As the Information Retrieval (IR) field increasingly recognizes the importance of inclusivity, addressing the needs of low-resource languages remains a significant challenge. This paper introduces the first large-scale Urdu IR dataset, created by translating the MS MARCO dataset through machine translation. We establish baseline results through zero-shot learning for IR in Urdu and subsequently apply the mMARCO multilingual IR methodology to this newly translated dataset. Our findings demonstrate that the fine-tuned model (Urdu-mT5-mMARCO) achieves a Mean Reciprocal Rank (MRR@10) of 0.247 and a Recall@10 of 0.439, representing significant improvements over zero-shot results and showing the potential for expanding IR access for Urdu speakers. By bridging access gaps for speakers of low-resource languages, this work not only advances multilingual IR research but also emphasizes the ethical and societal importance of inclusive IR technologies. This work provides valuable insights into the challenges and solutions for improving language representation and lays the groundwork for future research, especially in South Asian languages, which can benefit from the adaptable methods used in this study.

\keywords{Natural Language Processing  \and Information Retrieval \and Low-resource language.}

%% file: introduction.tex
Advancements in Natural Language Processing (NLP), particularly transformer architectures, have revolutionized Information Retrieval (IR), enhancing information access worldwide. However, this progress predominantly benefits high-resource languages, leaving low-resource languages like Urdu underrepresented. Urdu is spoken by over 70 million people primarily in South Asia, and it features unique linguistic characteristics, such as its Perso-Arabic script with right-to-left writing direction and rich morphology \cite{urdustats}, which pose significant challenges for IR systems. These include complex tokenization, script handling, and morphological analysis, which are also present in some high-resource languages such as Arabic, Chinese, and Russian but have benefited from extensive research and resource availability, making these challenges less of a barrier in those languages.

One of the primary challenges in expanding IR capabilities for low-resource languages like Urdu is the scarcity of large-scale, high-quality datasets necessary for training and evaluating models. Creating such datasets manually is resource-intensive and often impractical for languages with limited digital presence. Consequently, to close this gap, researchers have explored alternative methods, such as machine translation of existing datasets, but it introduces other challenges like translation errors and context loss, adversely affecting IR performance. To our knowledge, no prior work has specifically addressed these issues for IR in Urdu, leaving a gap that our work aims to fill.

In this paper, we address this issue by translating a large-scale MS MARCO passage ranking dataset \cite{msmarco} into Urdu using the IndicTrans2 model \cite{indictrans2}. This translation provides a valuable resource for Urdu IR development. We first established baseline performance on our newly created dataset using BM25 and mMarco (a multilingual mT5 model trained on the MS MARCO dataset)\cite{mmarco} in a zero-shot setting, as it was not originally trained on Urdu.  We then fine-tune the model on our Urdu-translated dataset. Our experiments demonstrate significant improvements in Mean Reciprocal Rank (MRR) and Recall metrics compared to the zero-shot results, highlighting the potential of our approach.

Our key contributions are:
\begin{itemize}
    \item Creation of the Urdu MS MARCO Dataset: We provide the first large-scale IR dataset for Urdu by translating the MS MARCO dataset using the IndicTrans2 model, addressing the data scarcity issue.
    \item Establishment of Baseline IR Performance: We conduct experiments using BM25 and a fine-tuned mT5 model to establish baseline retrieval performance on the Urdu dataset.
    \item Insights into Low-Resource Language IR: We analyze the challenges and potential solutions for improving IR in Urdu, providing a foundation for future research in similar languages.
    \item Publicly Available Resources: To facilitate future research in Urdu IR, we make our translated dataset, fine-tuned mT5 model, and code publicly available on Hugging Face \footnote{\url{https://huggingface.co/datasets/Mavkif/urdu-msmarco-dataset}} \footnote{\url{https://huggingface.co/Mavkif/urdu-mt5-mmarco}} and GitHub \footnote{\url{https://github.com/UmerTariq1/Urdu_MsMarco_Translation_Retrieval}}.
    
\end{itemize}

%% file: related_work.tex
The lack of large-scale, high-quality datasets for low-resource languages has been a significant barrier to advancing IR capabilities. To address this data scarcity, several efforts have created multilingual datasets through machine translation. For example, mMarco\cite{mmarco} extended the MS MARCO dataset\cite{msmarco} to 13 languages using Helsinki NLP models \cite{helsinki} and Google Translate. However, Urdu was not included, partly due to the lack of high-quality translation models for the language at that time.

Recent developments in machine translation, such as the IndicTrans2 model\cite{indictrans2}, have improved translation quality for South Asian languages, including Urdu. IndicTrans2 achieves a chrF++\footnote{Character-level F-score, used for capturing partial matches and morphological variations across languages} score of 68.2 for English-to-Urdu translation, surpassing models like Google Translate. This improvement opens new possibilities for creating large-scale Urdu datasets suitable for training IR models.

In the domain of multilingual NLP, models like mT5\cite{mt5} have demonstrated strong performance across multiple languages by pre-training on a massive multilingual corpus. Similarly, several works have focused on multilingual IR. ColBERT-X\cite{colbertx} extends the ColBERT\cite{colbert} architecture to multilingual settings, showing effectiveness in languages included in its training data. Benchmarks like MIRACL\cite{miracl} and initiatives like CIRAL\cite{ciral} aim to standardize evaluation and improve cross-lingual retrieval, particularly for low-resource scenarios. MIRACL addresses this by providing a diverse set of 18 languages, including both high and low-resource languages from 10 different language families, with a focus on monolingual retrieval. It features a variety of tasks, including question answering and document retrieval, and incorporates challenges like code-switching and dialectal variations. While MIRACL includes a diverse set of languages, it currently lacks a dedicated Urdu corpus.

Our approach specifically addresses this gap for Urdu, leveraging the improved translation capabilities of IndicTrans2 to create a large-scale Urdu IR dataset. By fine-tuning the mT5 model on this dataset, we address the challenges posed by Urdu's script and morphology. This work provides a foundation for future research in low-resource languages and contributes to the inclusivity of IR technologies.

%% file: dataset.tex
\subsection{Ms-Marco dataset}
The MS MARCO Passage ranking dataset \cite{msmarco} \footnote{https://microsoft.github.io/msmarco/Datasets.html} is a cornerstone in IR research, comprising over 8.8 million passages and more than 500 thousand unique queries each paired with at least one relevant passage. The commonly used development set includes 6,980 queries, while the training set is organized into 39 million triples, each consisting of a query, a relevant passage, and a non-relevant passage. An example of it can be seen in Table \ref{tab:dataset_example}. This dataset's scale and richness have significantly advanced English language IR models. However, replicating such progress for Urdu is challenging due to the absence of equivalent large-scale datasets or the prohibitive costs associated with translating such extensive corpora\footnote{estimated around \$30,000 for MS Marco, using services like Azure Translator or Google Translation API with their standard rates of 10-20\$ per million characters}.

\subsection{Translation Process}
To create the Urdu MS MARCO dataset, we translated the entire English MS MARCO dataset, including both passages and queries from the training and development sets, using the open-source IndicTrans2\cite{indictrans2} model. We utilized the distilled 200-million parameter version of IndicTrans2 to translate the entire MS MARCO dataset into Urdu. The process involved tokenizing the text with the IndicTransTokenizer, translating in batches to optimize computational efficiency and leveraging GPU acceleration. Key parameters included a batch size of 32 and a maximum sequence length of 512 tokens to handle lengthy passages. We also performed preprocessing steps, such as normalization and error correction to enhance the quality and usability of the dataset and the model. Overall the translation took around 120 hours to translate the dataset on a single V100 GPU with 32GB Vram. 

\begin{table}
    \caption{Example of a query and its relevant and non-relevant passage in both English and machine-translated Urdu. More examples can be seen on huggingface dataset webpage}
    \label{tab:dataset_example}
    \centering
\includegraphics[width=1\linewidth, trim={0cm 0cm 0cm 0cm}, clip]{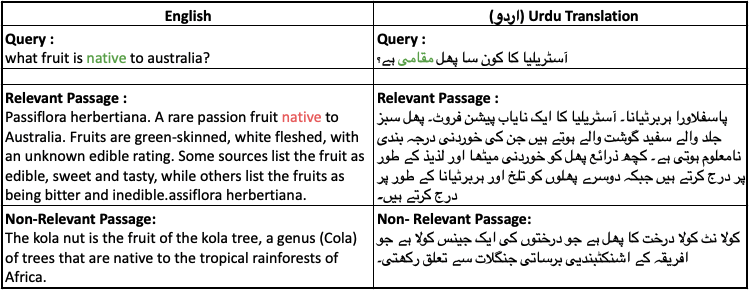}
\end{table}

%% file: setup.tex
\subsection{Baseline Models}
Given the absence of prior work on Urdu IR, establishing strong baseline performances is essential for meaningful comparison. We employed BM25, a well-established non-neural retrieval method \cite{bm25}, as our initial retrieval model to retrieve the top 1,000 passages per query\cite{irpipeline}. BM25 serves as a fundamental baseline due to its simplicity and effectiveness in various IR tasks. However, its performance tends to be limited for low-resource languages, as it does not account for complex morphological patterns or linguistic nuances present in a language.

To complement BM25, we also evaluated the zero-shot performance of the mMARCO model \cite{mmarco} as a re-ranker. Although Urdu is not included in mMARCO’s training data, we aimed to assess whether the model could offer effective re-ranking for Urdu text based solely on its multilingual training. This evaluation provides an initial benchmark for the model’s performance on low-resource languages like Urdu.

\subsection{Urdu Re-ranker Fine-tuned Model}
To enhance retrieval performance beyond the baseline, we fine-tuned the mMARCO model, a fine-tuned version of mT5, on our Urdu-translated MS MARCO dataset, following the original mMARCO training setup for consistency. We set the batch size to 32 with gradient accumulation over 4 steps, resulting in effective batch size of 128 to balance computational efficiency with model performance. The model used a constant learning rate of 0.001 and a dropout rate of 0.1 across all layers, while other hyperparameters remained aligned with the mMARCO setup.


In fine-tuning, we adapted our model for binary relevance classification, generating the urdu word for "yes" for relevant passages and the urdu word for "no" for non-relevant ones, thus framing relevance as a sequence-to-sequence task. This approach provides interpretable relevance decisions, with each query-passage pair yielding a clear binary response.


At inference time, we compute relevance scores by applying a softmax over the logits of the word "yes" and "no" tokens (in urdu script), ranking documents based on the probability of the urdu "yes" token. We fine-tuned the model using the full Urdu-translated MS MARCO dataset, consisting of 39 million triples (query, relevant, and non-relevant passages), with training taking approximately 75 hours on a single A100 GPU with 40GB Vram. This setup allowed the model to adapt effectively and make relevance predictions suited to its linguistic characteristics.

\subsection{Evaluation Metrics}

To assess the effectiveness of our models, we used common evaluation metrics in information retrieval research to ensure comparability with other studies. These metrics include Mean Reciprocal Rank at 10 (MRR@10), which evaluates the ranking quality by considering the position of the first relevant result within the top 10 retrieved passages. We also used Recall at 10 (Recall@10) to measure the proportion of relevant passages retrieved in the top 10 results, reflecting the system's ability to find as many relevant passages as possible. These two metrics were also used in the original mMarco benchmark. Additionally, we included Mean Average Precision at 10 (MAP@10) to assess the overall ranking quality across all relevant passages for each query. Finally, we used Normalized Discounted Cumulative Gain at 10 (NDCG@10), a more sophisticated metric that takes into account the position of all relevant passages in the ranking and gives higher weight to those appearing earlier in the list. By employing these metrics, we aimed to provide a comprehensive evaluation of retrieval performance, encompassing various aspects of ranking quality and relevance.


\vspace{-05pt}
\begin{table}
\caption{Performance Comparison of Retrieval Models on English and Urdu MS-Marco Datasets}
\label{results}
\begin{adjustbox}{center} 
\begin{tabular}{|l|l|l|l|l|l|} 
\hline
{\bfseries Model } &  {\bfseries  Dataset Language } & {\bfseries   Recall@10  } & {\bfseries   MRR@10  } &  {\bfseries   MAP@10  } & {\bfseries   NDCG@10  }\\
\hline
BM25 (k=1000)  &  English & 0.391 & 0.187 & 0.104 & 0.275 \\
mMARCO Reranker  &  English & 0.639 & 0.370 & 0.245 & 0.498 \\
\hline
BM25 (k=1000)  &  Urdu & 0.247 &  0.121  & 0.070 & 0.177 \\
Zero-Shot mMARCO Reranker &  Urdu & 0.408  & 0.204 & - & - \\
Urdu mT5-mMARCO ( {\bfseries This Work} ) &  Urdu & {\bfseries 0.438}  & {\bfseries 0.248} & {\bfseries 0.159} & {\bfseries 0.340} \\
\hline
\end{tabular}
\end{adjustbox} 
\end{table}

\vspace{-05pt}

%% file: results.tex
\subsection{Performance Comparison}

We evaluated the retrieval performance on the Urdu dataset using BM25 for initial retrieval, followed by different re-ranking strategies (see Table \ref{results}). The baseline BM25 model is used to retrieve the top 1000 passages per query and it achieves an MRR@10 of \textbf{0.121} on the Urdu dataset. The baseline BM25 model exhibited significantly lower performance on the Urdu dataset compared to the English dataset. This disparity underscores the challenges of applying traditional retrieval techniques to morphologically rich languages like Urdu, where effective tokenization requires more nuanced approaches than those typically used for English.  The limitations of standard tokenizers for handling Urdu morphology likely contribute to the lower performance observed with BM25.


Applying the zero-shot mMARCO re-ranker to the BM25 outputs improved performance, despite the model not being trained on Urdu. The mMARCO model generalized to Urdu to some extent, likely due to its multilingual training on languages with similar scripts, such as Arabic. This suggests that the model's semantic understanding and script familiarity enabled it to enhance the ranking of relevant passages over the BM25 baseline.


Our fine-tuned model, Urdu mT5-mMARCO, yielded the most substantial gains. Fine-tuning on the Urdu dataset allowed the model to adapt to the nuances of the language, leading to improvements across all reported metrics. Notably, Urdu mT5-mMARCO achieved a Recall@10 of \textbf{0.438}, surpassing both the BM25 baseline (0.247) and the zero-shot mMARCO (0.408). Furthermore, the fine-tuned model exhibited improvements in MRR@10 (\textbf{0.248}), exceeding the zero-shot model (0.204) and significantly outperforming the BM25 baseline (0.121). The gains observed in MAP@10 (\textbf{0.159}) and NDCG@10 (\textbf{0.340}) further demonstrate the positive impact of fine-tuning. This highlights the effectiveness of language-specific fine-tuning, even with machine-translated training data, in enhancing retrieval performance for low-resource languages. The results demonstrate the potential of this approach to bridge the gap in information access for Urdu speakers.

\vspace{-5pt}
\subsection{Impact of Translation Quality and Limitations}
We acknowledge that machine translations, like those from IndicTrans2, may not match the precision of human annotations, and translation errors can misalign query and passage semantics, impacting retrieval precision. For example, in table \ref{tab:dataset_example}, the word \textit{native} is correctly translated in the query (highlighted in green) but omitted in the relevant passage translation (highlighted in red) by the model. Nonetheless, this work serves as a crucial first step toward enabling Urdu IR research. Furthermore, open-sourcing the translation pipeline, data, and model 
aims to support future multilingual IR projects and address data limitations in low-resource languages beyond Urdu, fostering broader inclusivity in IR research.

\vspace{-5pt}

%% file: conclusion.tex
This paper introduces a foundational approach for enabling Information Retrieval (IR) in low-resource languages by translating the MS MARCO dataset into Urdu and fine-tuning the mMARCO model on this data. The baseline BM25 model for Urdu showed significantly lower performance compared to English, underscoring the need for tailored models in Urdu IR. Our results demonstrate that fine-tuning significantly enhances retrieval performance, highlighting the potential of machine translation as a first step in overcoming data limitations for low-resource languages. 

This work contributes to more inclusive IR by making retrieval systems accessible to Urdu speakers, with implications for other South Asian languages. Future work could involve incorporating manual verification to refine translation quality and exploring hybrid methods that combine machine and human translations. To better understand the impact of these methods, a comprehensive analysis of translation quality and its quantitative effect on retrieval performance will be conducted. Expanding this approach to other low-resource languages would further support the development of fair, balanced and multilingual IR technologies.

%% file: acknowledgement.tex
This research was conducted at the German Research Center for Artificial Intelligence (DFKI) and the University of Saarland. The authors gratefully acknowledge DFKI for providing the necessary hardware resources and the supportive research environment. This work was also supported by the German Federal Ministry of Education and Research (BMBF) as part of the TRAILS project (01IW24005). The authors thank the BMBF for their commitment to advancing research in Artificial Intelligence.